# Monitoring Urban Traffic Dynamics at High Spatiotemporal Resolution Using Distributed Acoustic Sensing and Deep Learning

Hao Tian[a], Heng Cai[a*], Xiaowei Chen[b], Yifan Yang[a]

[a]*Department of Geography, Texas A&M University, College Station, TX 77840, USA;*

[b]*Department of Geology and Geophysics, Texas A&M University, College Station, TX 77840, USA.*

* Corresponding author: hengcai@tamu.edu

Mapping the distribution of traffic dynamics at high spatiotemporal resolution is a fundamental question in transportation research. Distributed acoustic sensing (DAS), an innovative seismic observation tool, emerges as a promising solution for real-time urban traffic monitoring at high spatial and temporal scales. Distributed acoustic sensing repurposes existing underground fiber-optic cables as dense, continuous sensor arrays, enabling passive and privacy-preserving monitoring of roadway traffic activity at meter-level spatial and second-level temporal resolution. This study examines whether integrating DAS and deep learning models can serve as a continuous and efficient urban traffic observatory for revealing urban traffic dynamics (i.e. traffic volume and congestion, event-driven changes) at high spatiotemporal resolution. Using a DAS deployment along a roadway network in the City of College Station, Texas, USA, this study develops a deep learning-empowered analytical framework that converts raw ground vibration waveforms into spatiotemporal representations, detects vehicle trajectory, and infers traffic states from aggregated traffic volume and speed. A hybrid training strategy combining synthetic and manually annotated DAS images is used to improve vehicle detection under noisy and congested conditions, with model outputs further aggregated to characterize system-level traffic dynamics. The results show that different urban events generate distinct traffic-state signatures. Home football gamedays produce localized congestion amplification with multi-peaked surges associated with event-related arrival and departure patterns, whereas the Thanksgiving holiday leads to sustained traffic suppression and flattened diurnal traffic rhythms, alongside substantial spatial heterogeneity. The integration of distributed acoustic sensing and deep learning provides a promising urban traffic sensing approach for monitoring traffic dynamics at high spatiotemporal resolution with the potential for real-time and computationally efficient analysis. The findings highlight the potential of DAS to support efficient, accurate, and privacy-preserving traffic monitoring, particularly when conventional traffic sensing data are sparse, unavailable, or disrupted.

## 1. Introduction

Transportation dynamics represents one of the most direct and continuously observable proxies of collective human activity in urban environments. Real-time and accurate monitoring of transportation dynamics not only aids in assessing roadway service levels but also plays a critical role in understanding environmental exposure, emergency response efficiency, and broader urban resilience (Brent & Beland, 2020). Consequently, transportation monitoring has become a fundamental component of the real-time urban observation framework that supports smart city development and resilient governance (Umeike et al., 2025).

Recent advances in urban sensing technologies and big data analytics have enabled a wide range of traffic monitoring approaches, including cameras, radar, infrared sensors, and vehicle-based GPS data (Chen et al., 2021; Herrera et al., 2010; Zhan et al., 2020). While these data sources provide high spatial and temporal resolution under normal operating conditions, their practical deployment is often constrained by privacy concerns, uneven or sparse sampling coverage, and operational vulnerability (Thabit et al., 2024). During disruptive events such as hurricanes, floods, or widespread power outages, many of these sensing systems become disrupted or entirely unavailable, limiting their effectiveness when timely situational awareness is most needed (Guerrero-Ibáñez et al., 2018). These limitations motivate the exploration of complementary sensing modalities that are scalable, resilient, and privacy-preserving (Tian et al., 2026).

Distributed Acoustic Sensing (DAS) has recently emerged as a promising alternative for urban transportation monitoring under such constraints (Min et al., 2024; Z. Wang et al., 2025; Xie et al., 2025; Yuan et al., 2024). DAS is a sensing technology that turns a fiber-optic cable into a dense, continuous array of vibration sensors. Instead of deploying discrete instruments, DAS measures signals along the entire length of the cable (often tens of kilometers), producing high-resolution spatiotemporal data capturing the ground vibration caused by various sources. It operates by launching laser pulses into standard fiber-optic cables and analyzing the intensity and phase of Rayleigh backscattered light, thereby enabling the reconstruction of ground vibrations and disturbances along the cable with meter-scale spatial resolution (Lindsey & Martin, 2021; Zhan, 2019). The unique advantage of DAS is the ability to transform existing urban communication fiber networks into dense sensor arrays deployed along roadways. Compared to traditional node-based sensor systems, DAS offers broader spatial coverage, lower investment and maintenance costs, faster scalability, and passive operation that avoids the collection of personally identifiable information. These advantages have also facilitated the application of DAS across multiple domains, including near-surface imaging (Ajo-Franklin et al., 2019; Lai et al., 2024), critical infrastructure monitoring (Cheng et al., 2023), and human locomotion classification (Peng et al., 2020).

Given the widespread availability of fiber-optic infrastructure in urban areas, much of which is deployed along roadways, traffic-related signals are among the most frequently captured by DAS, making it a particularly attractive modality for transportation monitoring (Anderson et al., 2024). In recent years, a growing body of literature has highlighted the broad potential of DAS in transportation, offering valuable insights into vehicle detection and classification, traffic flow and speed estimation, as well as roadway condition monitoring (Chambers, 2020; Liu et al., 2020; Wang et al., 2021). Although these signal-processing-based approaches have proven effective, they typically rely on complex handcrafted pipelines and can be computationally intensive, limiting their scalability for real-time, large-scale applications. Recent advances in machine learning and deep learning provide new opportunities for DAS-based transportation monitoring (Min et al., 2024; Z. Wang et al., 2025). Unlike traditional signal-processing approaches that rely on handcrafted features, machine learning or deep learning methods can automatically learn discriminative features from large-scale DAS datasets, thereby enhancing processing speed, scalability, and real-time applicability (Ye et al., 2023). These advances have substantially improved the capability of DAS for vehicle detection and trajectory extraction.

Despite these important advances, several critical gaps remain from the perspective of transportation geography and urban traffic monitoring. First, most existing studies remain largely focused on vehicle-level detection, trajectory extraction, and speed estimation, with far less attention devoted to translating these observations into higher-level traffic-state indicators for transportation-system-level analysis. Second, although DAS has demonstrated considerable potential for continuous traffic monitoring, its capability to characterize how different disruptive events reshape urban traffic dynamics over time and across roadway networks remains largely unexplored. Such event-scale analyses are essential for understanding traffic resilience, non-recurring congestion, and the spatial heterogeneity of urban transportation dynamics. Third, despite the emergence of deep learning methods for DAS-based traffic monitoring, current approaches generally rely on limited labeled datasets and often exhibit reduced robustness under noisy and congested traffic conditions, thereby limiting their scalability for continuous urban traffic monitoring. Addressing these challenges therefore requires not only more robust traffic inference methods, but also analytical frameworks capable of translating vehicle-level observations into transportation-system-level understanding of urban traffic dynamics and resilience.

To bridge these gaps, this study investigates whether DAS can serve as a continuous urban traffic observatory for monitoring event-driven traffic dynamics at high spatiotemporal resolution and for advancing transportation-system-level understanding of urban mobility. To achieve this objective, we develop a deep learning-enabled analytical framework that integrates a hybrid synthetic-real training strategy with traffic-state inference from DAS observations, enabling robust vehicle detection and the aggregation of vehicle-level information into system-level traffic indicators.

Unlike previous DAS studies that primarily emphasize vehicle detection and trajectory extraction, the novelty of this study lies in extending DAS-based sensing toward transportation-system-level understanding through traffic-state inference and event-driven traffic analysis. This study makes three primary contributions. First, we move beyond conventional vehicle-level DAS applications by developing a framework that translates vehicle-level observations into transportation-system-level traffic-state indicators for continuous urban traffic monitoring. Second, we develop a hybrid synthetic-real training strategy that substantially reduces dependence on manually labeled datasets while maintaining robust traffic inference under noisy and congested conditions. Finally, we demonstrate how continuous DAS observations can characterize event-driven traffic dynamics under different disruptive events, revealing distinct patterns of temporal mobility reorganization and spatial heterogeneity in traffic responses across roadway segments.

Guided by these objectives, this study addresses three interrelated research questions: (1) Can a deep learning-enabled DAS analytical framework support robust and computationally efficient inference of urban traffic states at a high spatiotemporal resolution? (2) How do different types of disruptive events reshape urban traffic dynamics over time? (3) To what extent do event-driven traffic responses vary spatially across roadway segments?

## 2. Literature Review

### *2.1 Distributed acoustic sensing (DAS) as an emerging urban sensing infrastructure*

DAS is a fiber-optic sensing technology that measures dynamic strain along the length of an optical fiber by analyzing Rayleigh backscattered light from coherent laser pulses (A. H. Hartog, 2017). By leveraging phase changes in the backscatter signal, DAS can detect vibrations with meter-scale spatial resolution and millisecond-scale temporal resolution over distances spanning tens of kilometers (Zhan, 2019). This capability transforms existing telecommunication fiber-optic cables, including unused dark fibers, into dense, continuous sensor arrays without the need for additional hardware deployment at each sensing point (Ajo-Franklin et al., 2019).

The technology was initially developed for applications in oil and gas exploration (Mateeva et al., 2014). Over the past decade, DAS has rapidly expanded into diverse geoscience and engineering domains. In seismology, DAS arrays have been used for ambient noise tomography and earthquake detection in both urban and submarine environments (Lindsey & Martin, 2021; Sladen et al., 2019). For infrastructure monitoring, DAS has been applied to detect structural anomalies in railways, pipelines, and bridges (Hussels et al., 2019; Zhong et al., 2026). In urban settings, Ajo-Franklin et al. (2019) showed that urban-scale dark fiber can support near-surface characterization and broadband seismic event detection, while Lai et al. (2024) extended passive seismic imaging with DAS to urban settings, producing shear-wave velocity models beneath Melbourne, Australia. More recently, Liu et al. (2025) leveraged existing telecommunication fiber-optic networks to enable city-scale seismic source mapping, demonstrating the potential of DAS for urban sensing. Taken together, these developments highlight DAS as a dense, continuous, and scalable sensing infrastructure, providing a foundation for emerging applications in urban system monitoring beyond traditional geophysical domains.

### *2.2 DAS-based traffic monitoring using signal-processing methods*

Given the widespread co-location of fiber-optic cables and roadways, traffic-induced ground vibrations are among the most prominent signals captured by urban DAS deployments. The earliest transportation applications of DAS were dominated by physics-based signal-processing workflows. Liu et al. (2020) pioneeringly employed an improved wavelet-denoising and dual-threshold algorithm to process DAS signals for vehicle count and speed estimation. Their approach demonstrated that individual vehicle passages could be reliably identified from the temporal patterns of DAS strain-rate signals. Building on this foundation, Lindsey et al. (2020) developed a template-matching algorithm to detect vehicles from low-frequency DAS strain signals recorded on dark fiber in Stanford, California. Their study provided one of the first demonstrations of using DAS to monitor traffic response patterns during the COVID-19 pandemic lockdowns, revealing significant reductions in traffic volume and increases in average speed during shelter-in-place orders. Wang et al. (2021) further showed that vehicle volume and mean speeds can be extracted from DAS data by applying a 4th-root slant stacking approach, enabling city-scale monitoring of traffic responses to COVID-19 lockdowns.

However, the same literature also exposes the limitations of classical signal-processing approaches. First, the computational cost can become prohibitive as fiber length and monitoring duration increase, limiting its use in real-time applications (Deng et al., 2025). Second, their performance tends to degrade under congested conditions where vehicle-induced wavefields overlap spatially and temporally (Chambers, 2020). For this reason, signal-processing methods are not always well matched to large-scale, near-real-time monitoring under noisy urban conditions.

### *2.3 DAS-based traffic monitoring using deep learning methods*

Recent advances in machine learning and deep learning have opened new opportunities for DAS-based traffic monitoring by enabling automated feature extraction from large-scale spatiotemporal data. Unlike traditional signal processing approaches that rely on handcrafted features, deep learning methods can automatically learn discriminative features from raw or minimally processed DAS data, thereby enhancing processing speed, scalability, and real-time applicability (Z. Wang et al., 2025; Ye et al., 2023).

Initial efforts in applying machine learning to DAS traffic data focused on traditional classification models. For example, Min et al. (2024) applied a series of preprocessing filters to enhance vehicle signals from raw DAS recordings, followed by a support vector machine classifier to distinguish traffic-related signals from background noise. While effective for binary classification tasks, SVM-based approaches do not readily generalize multi-vehicle detection and localization in spatiotemporal data. In addition, increasing research has explored converting raw DAS data into spatiotemporal images, where vehicle trajectories appear as coherent linear features. This representation enhances the visual distinctiveness of vehicle-induced signals, allowing deep learning models to be easily applied for accurate vehicle detection, trajectory tracking, and speed estimation. Ye et al. (2023) were among the first to apply a YOLO-based (You Only Look Once) object detection framework to DAS spatiotemporal images for real-time vehicle detection. In parallel, transformer-based detection architectures have emerged as competitive alternatives to convolutional detectors. RT-DETR (Real-Time DEtection TRansformer), introduced by Zhao et al. (2024), represents the first end-to-end transformer-based detector designed explicitly for real-time applications. However, the relative performance of transformer-based versus convolutional detectors for DAS spatiotemporal images remains largely unexplored.

Beyond standard object detection, several specialized deep learning approaches have been proposed for DAS traffic analysis. Ende et al. (2023) introduced a deconvolution autoencoder for traffic analysis, trained to remove the spatially extended impulse response of the DAS system, and thereby sharpen individual vehicle signatures. Yuan et al. (2024) developed a spatial U-Net architecture incorporating physics-based wavelet representations, which improved trajectory extraction accuracy by leveraging domain-specific knowledge about wave propagation characteristics. Xie et al. (2025) introduced a CNN–Hough transform approach to robustly extracting vehicle trajectories in noisy conditions.

Despite these advances, most existing studies primarily focus on vehicle-level detection and trajectory extraction, while their potential for large-scale traffic state inference and event-level analysis remains largely underexplored. Furthermore, the scalability of deep learning-based DAS applications is constrained by the limited availability of manually annotated datasets, particularly for continuous monitoring under diverse traffic conditions. Addressing these analytical and methodological challenges requires moving beyond vehicle-level sensing toward scalable transportation-system-level traffic analysis.

### *2.4 Event-driven traffic dynamics and urban resilience*

Understanding how disruptive events reshape urban traffic patterns is a central concern in transportation planning, operations, and resilience assessment. Planned special events such as sporting competitions and concerts generate significant non-recurring congestion, with traffic volume increases of 10% to 50% depending on event size, venue location, and the capacity of the surrounding road network (Giuliano & Lu, 2021; Kwoczek et al., 2014). Importantly, these demand surges interact with the underlying road network structure, leading to uneven traffic responses across different locations. This spatial variability has been empirically documented in prior studies. Donovan and Work (2017) used large-scale taxi GPS trajectory data to analyze spatiotemporal traffic dynamics in New York City during Hurricane Sandy, revealing substantial variability in traffic conditions across both space and time, with clear differences associated with roadway function and network location. This observed heterogeneity has motivated subsequent efforts to explicitly characterize the structural and functional importance of road segments. In this context, Feng et al. (2025) also used taxi GPS trajectory data from Beijing's inner 4th Ring Road under adverse weather conditions to evaluate road segment criticality and showed that urban road network resilience is spatially differentiated, with high-criticality segments concentrated on major arterials and key intersections rather than being uniformly distributed across the network.

Despite these advances, most event-driven traffic analyses have relied on conventional sensing modalities, including loop detectors, GPS-based trajectory data, and camera-based observations, which are subject to the coverage, privacy, and resilience limitations discussed previously. The potential of DAS to characterize event-driven traffic dynamics at fine spatiotemporal resolution has been suggested by recent studies but has not been systematically explored (Tian et al., 2026). The continuous, privacy-preserving, and infrastructure-based nature of DAS systems provides a unique opportunity to examine how different event types produce distinct traffic-state signatures and how these responses vary spatially across roadway segments.

## 3 Data and methods

### *3.1 Data collection*

This study utilizes Distributed Acoustic Sensing (DAS) data obtained from a fiber-optic cable deployed on the Texas A&M University (TAMU) campus in College Station, Texas. The cable is connected to a DAS interrogator unit, which converts the fiber into a dense array of virtual sensors that continuously record spatiotemporal strain (vibration) signals along its length (Fig. 1a). The monitored fiber in College Station, TX spans approximately 23 km and follows roadways, campus facilities, and rail-adjacent corridors, allowing the capture of diverse traffic conditions within a compact urban environment. The fiber-optic cable was connected to a Silixa iDAS V2.5 interrogator, configured with a gauge length of 10 m, a channel spacing of 4 m, and a sampling rate of 100 Hz. The geographic locations of the DAS channels were determined using a GPS-tracked vehicle following the geolocation procedure of Biondi et al. (2022). The resulting channel mapping was subsequently verified using controlled hammer (tap) tests at several known locations. Under this configuration, the DAS system transforms the telecommunication fiber network into thousands of spatially contiguous sensing channels with meter-scale spatial resolution and continuous temporal coverage. The system records axial strain-rate measurements, which capture ground vibrations induced by passing vehicles, trains, and other anthropogenic activities along the fiber route (Fig. 1b). With this setup, the system operated continuously throughout October and November 2023 and generated approximately 6.2 TB of raw strain-rate measurements.

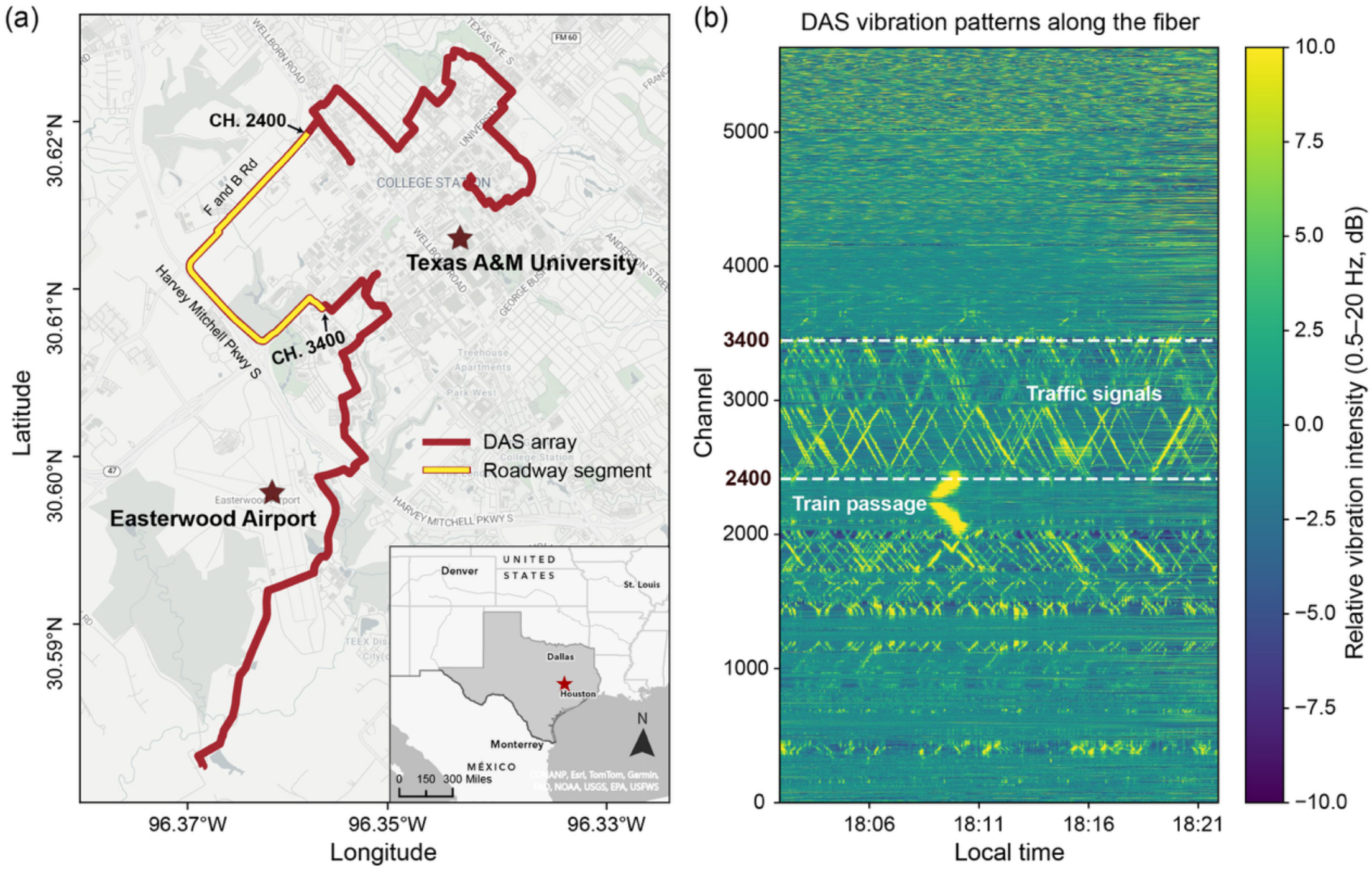

Fig. 1. Study area and DAS ground vibration observations along the Texas A&M University campus fiber. (a) Layout of the DAS array and selected roadway segments (highlighted in yellow). (b) Waterfall plot of band-limited vibration intensity (0.5–20 Hz), highlighting traffic-related signals and a train passage. In the waterfall plot, each horizontal row corresponds to a DAS channel, which represents a virtual sensor at a specific location along the fiber-optic cable (spaced 4 m apart in this deployment). The horizontal axis denotes time, and the vertical axis denotes channel number (i.e., spatial position along the fiber). Pixel intensity reflects the amplitude of ground vibration energy recorded at each channel and time step.

### *3.2 Data preprocessing and spatiotemporal representation*

Raw DAS recordings consist of continuous strain-rate measurements along the fiber, representing a superposition of anthropogenic signals (e.g., vehicle-induced vibrations), ambient environmental noise, and system-related noise, including instrumental noise and variability in fiber–ground coupling. To enhance traffic-related signals and construct representations suitable for both deep learning and slant-stacking algorithms, a multi-stage preprocessing pipeline was developed (Fig. 2). We elaborate on each stage in Section 3.2, 3.3, and 3.4. Here, slant-stacking refers to a classical signal-processing technique originally developed in seismology that enhances coherent linear features in space–time data by summing (stacking) signals along candidate slopes. Applied to DAS waterfall images, it emphasizes the linear trajectories left by moving vehicles and has been used as a baseline for DAS-based traffic monitoring (Wang et al., 2021)

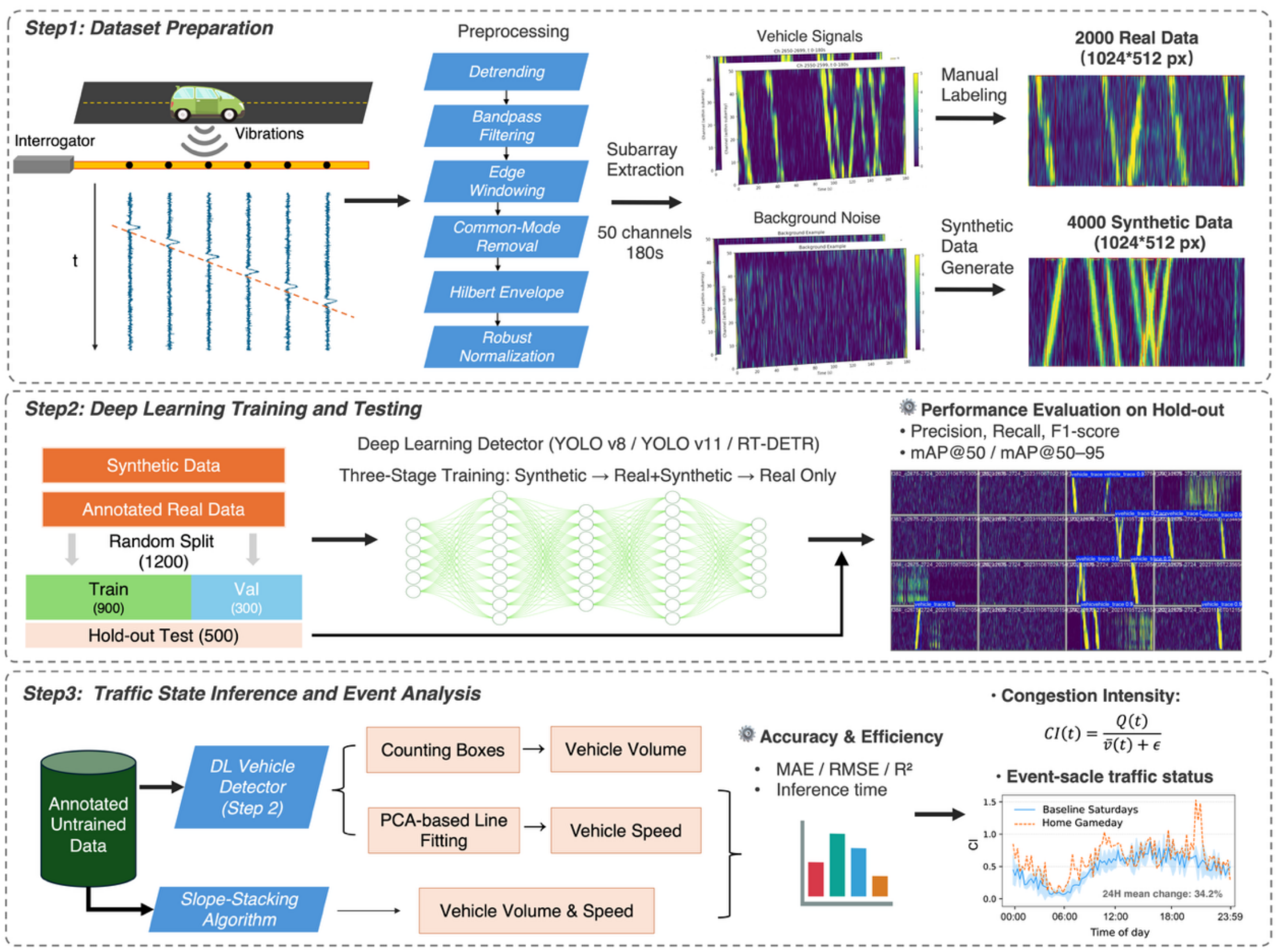

Fig. 2. Overview of the proposed DAS-based analytical framework for urban traffic state inference, including dataset preparation, deep learning-based vehicle detection, and congestion-based event-scale traffic analysis.

First, the raw waveforms were detrended and band-pass filtered between 0.5 and 20 Hz to isolate frequency components predominantly associated with vehicle-induced ground vibrations. The selected 0.5-20 Hz band captures the frequency range where vehicle-induced vibrations exhibit the highest temporal coherence and signal-to-noise contrast relative to background fluctuations (Fig. S1). Additional preprocessing steps, including edge tapering, common-mode noise suppression, envelope extraction, smoothing, and robust normalization, were applied to further enhance coherent traffic-related features while suppressing residual noise and outliers (Fig. S2). Specifically, edge tapering was used to reduce boundary artifacts introduced during signal segmentation; common-mode noise removal suppressed spatially coherent background noise shared across DAS channels, such as instrumental and environmental noise; envelope extraction emphasized the amplitude of vehicle-induced vibration signals while reducing sensitivity to rapid waveform oscillations; smoothing improved the continuity of vehicle trajectories by suppressing high-frequency fluctuations; and robust normalization scaled the signal amplitudes using statistics that are less sensitive to outliers, thereby improving the consistency of spatiotemporal representations for subsequent deep learning and slant-stacking analyses.

After preprocessing, the continuous DAS recordings were segmented into non-overlapping spatiotemporal subarrays with a fixed duration of 180s and a spatial extent of 50 channels (approximately 200m along the fiber). Each subarray was transformed into a two-dimensional spatiotemporal image, where the horizontal axis represents time, the vertical axis represents channels, and pixel values represent the normalized energy of DAS signals. In this space-time representation, moving vehicles manifest as coherent linear features, with their slopes indicating apparent velocities, thereby supporting both intuitive visualization and automated extraction of vehicle trajectories of traffic dynamics. It should be noted that the 4-m channel spacing defines the native sensing resolution at which individual vehicle trajectories are detected and their apparent speeds estimated, whereas the 50-channel (≈200 m) subarray defines the spatial window over which vehicle-level detections are subsequently aggregated into system-level traffic-state indicators. This window size was selected to ensure a sufficient number of vehicle detections within each analysis window for stable traffic-state estimation and thus represents a deliberate trade-off between spatial specificity and the statistical reliability of the resulting indicators.

To address the scarcity of labeled DAS traffic data, two complementary datasets were constructed from these spatiotemporal images. A real dataset consisting of 1,700 annotated images was created by visually identifying vehicle trajectories and labeling them with bounding boxes manually. In parallel, a synthetic dataset of 4,000 images was generated by embedding simulated vehicle trajectories into background noise extracted from real DAS recordings. Specifically, the synthetic generation process consisted of two components: (1) background noise extraction and (2) vehicle trajectory simulation. Background noise panels were obtained from late-night DAS recordings, when traffic activity was minimal. Candidate time windows were then manually inspected, and only those containing no identifiable coherent vehicle trajectories were retained, thereby preserving the authentic noise characteristics of the sensing environment (e.g., ambient vibrations and fiber-ground coupling variability). Vehicle-like signal patterns were then generated by constructing linear energy features in the space–time domain with parameterized slopes (corresponding to apparent vehicle speeds), amplitudes, and widths. These synthetic trajectories were designed to mimic the spatiotemporal signatures of real vehicle passages, with speeds uniformly sampled between 20 and 70 mph, and multiple trajectories superimposed to simulate multi-vehicle overlap scenarios. The simulated vehicle trajectories were additively embedded into the extracted noise panels, producing composite spatiotemporal images that closely resemble real DAS observations while providing precise ground-truth bounding box annotations. This approach preserves realistic noise characteristics while allowing controlled variation in apparent vehicle speed, congestion, and overlap conditions.

### *3.3 Traffic detection using deep learning models*

Traffic detection was formulated as a supervised object detection problem on preprocessed DAS spatiotemporal images. A total of 1,200 real labeled DAS images were randomly divided into training and validation sets with a 3:1 split. An additional 500 real images were reserved as an independent hold-out dataset and excluded from all training and validation procedures to enable an unbiased final performance evaluation. All manual annotations were generated by a single trained annotator following a consistent annotation protocol to maintain labeling consistency throughout the dataset. This resulted in 900 real images for training and 300 for validation. The 4,000 synthetic images were used exclusively in Stages 1 and 2 of the training strategy (described below), while the 900 real training images were introduced in Stage 2 alongside the synthetic data and used alone in Stage 3. The validation set was used consistently across all stages to monitor training convergence and select the best-performing model, whereas the 500-image independent hold-out dataset remained entirely unseen until the final performance evaluation.

To address the limited availability of labeled real DAS data and enhance robustness under diverse traffic conditions, a three-stage training strategy was adopted. This strategy draws on the principles of curriculum learning and transfer learning, where models are first exposed to large-scale, lower-quality data to learn generalizable representations before being progressively adapted to smaller, higher-quality datasets (Bengio et al., 2009; Zoph et al., 2020). By structuring the training pipeline as a progression from synthetic-only to hybrid to real-only data, each stage serves a distinct and complementary role in building detection capability. In the first stage 'Synthetic pre-training', detection models were trained exclusively on 4,000 synthetically generated DAS images to capture generalized spatiotemporal representations of vehicle-induced trajectories. The model can effectively learn fundamental geometric priors of vehicle trajectories, such as orientation, spatial continuity, and multi-object overlap patterns, without being constrained by the limited size of the real labeled dataset. In the second stage 'hybrid training', the pretrained models were fine-tuned using a hybrid dataset consisting of the synthetic dataset and real training dataset, enabling adaptation to realistic noise conditions while maintaining variability introduced by synthetic samples. This stage is critical because it bridges the domain gap between synthetic and real data: the model is recalibrated to adapt to the texture, noise structure, and amplitude characteristics unique to real DAS recordings, while retaining the broad feature representations acquired in the first stage. In the third stage 'real-only training', the models were further fine-tuned exclusively on the real training dataset to improve detection performance under real-world DAS conditions. This stage focuses on the model on the target data distribution, enabling it to better adapt to real signal characteristics.

Collectively, this three-stage progression follows a coarse-to-fine learning trajectory: The first stage establishes broad geometric representations, the second stage aligns these representations with real-world signal characteristics, and the third stage refines the detector for real-world data distribution. For comparison, an additional baseline model was trained using only the 900 real labeled DAS images without synthetic pretraining.

Three representative object detection architectures were evaluated, including YOLOv8, YOLOv11 and RT-DETR. YOLOv8 and YOLOv11 are one-stage convolutional detectors that perform object localization and classification in a single forward pass, enabling efficient real-time inference. As successive versions of the YOLO (You Only Look Once) family, they adopt anchor-free detection heads, enhanced feature pyramid backbones, and improved loss functions to better handle small and irregular targets (Redmon et al., 2016). RT-DETR, in contrast, is a transformer-based detector that eliminates non-maximum suppression through a set-based, query-driven decoding framework and captures long-range spatial dependencies via global self-attention (Zhao et al., 2024).

Model performance was assessed on an independent hold-out dataset using standard object detection metrics, including precision, recall, mAP@0.5, and mAP@0.5–0.95. Precision measures the proportion of correctly detected vehicle trajectories among all detections, while recall quantifies the proportion of true trajectories that are successfully identified. mAP@0.5 and mAP@0.5–0.95 summarize overall detection performance by integrating precision–recall trade-offs across different confidence thresholds and intersection-over-union (IoU) criteria, with mAP@0.5 using a fixed IoU threshold of 0.5 and mAP@0.5–0.95 averaging performance over multiple, increasingly strict IoU thresholds, thus providing a more comprehensive and stringent evaluation. All experiments were conducted on a Dell Precision 3660 workstation equipped with an Intel Core i9-12900K CPU, 64 GB RAM, and an NVIDIA GeForce RTX 3080 GPU.

### *3.4 Traffic activity inference and event-based comparison*

The vehicle trajectories detected by deep learning models provide vehicle-level information that can be further aggregated to infer traffic activity. For each detected vehicle, high-energy pixels associated with the corresponding bounding box were extracted in the time-channel domain to represent its spatiotemporal trajectory. Principal component analysis (PCA) was applied to these pixels to obtain a robust estimate of the dominant trajectory orientation (Jolliffe & Cadima, 2016). Within each detected vehicle segment, the trajectory was locally approximated as a linear structure over a short temporal window. This local approximation captures the dominant spatiotemporal evolution of vehicle-induced DAS signals while allowing for potential variations in vehicle motion over longer travel distances. The DAS-derived apparent vehicle speed $v_i$ was then estimated from the ratio between the local spatial displacement ($\Delta x_i$) and the corresponding temporal duration ($\Delta t_i$) of the first principal component trajectory,

$$v_i = \frac{\Delta x_i}{\Delta t_i} \tag{1}$$

$$\Delta x_i = \Delta c_i \cdot \Delta d \tag{2}$$

where $\Delta c_i$ represents the number of DAS channels spanned by the first principal trajectory, $\Delta d$ is the channel spacing (4 m in this study), and $\Delta t_i$ represents the corresponding temporal duration. The estimated velocity represents the apparent motion speed inferred from the local trajectory slope in DAS spatiotemporal representations. Compared with endpoint-based slope estimation, the PCA-based approach provides a more stable estimate of the apparent trajectory speed under noisy conditions and partial trajectory detections.

Traffic volume was estimated by counting the number of detected vehicle trajectories within each analysis window, while average apparent traffic speed was obtained by aggregating individual apparent vehicle speed estimates. In parallel, a classical 4th-root slant stacking approach was applied directly to the DAS wavefield to derive apparent vehicle speed and traffic activity indicators, serving as a signal-processing baseline for comparison with the deep learning-based approach. The 4th-root slant stacking approach is a signal processing method used specifically in geophysics and seismic data processing to enhance coherent linear features (e.g., vehicle trajectories) in space–time data while suppressing noise. Following the slant-stacking implementation of Wang et al. (2021), vehicle trajectories were identified using an adaptive peak-detection threshold. Specifically, for each spatiotemporal window, the threshold was determined as the stacking-energy value corresponding to the 68th percentile of the cumulative distribution function (CDF) of the stacking energy, with a minimum threshold of 2.5 to avoid unrealistically low threshold values. The inferred vehicle-level measurements were then temporally aggregated to characterize traffic activity patterns.

To characterize traffic conditions at the system level, detected vehicle activity and average speed were combined to construct a line-based congestion intensity index. For each analysis window $t$, congestion intensity was defined as

$$CI(t) = \frac{Q(t)}{\bar{v}(t) + \epsilon} \tag{3}$$

where $Q(t)$ denotes the number of detected vehicles within the window, $\bar{v}(t)$ represents the corresponding average apparent vehicle speed, and $\epsilon$ is a small constant introduced to avoid numerical instability under near-zero speed conditions. This formulation is motivated by the well-established interaction among traffic accumulation, speed reduction, and congestion formation in traffic flow theory (Treiber & Kesting, 2013). However, given that DAS observations provide aggregate vehicle signatures rather than direct measurements of lane-level flow and road occupancy, the proposed metric should not be interpreted as physical traffic density. Instead, we define it as a congestion intensity indicator that integrates two key characteristics of traffic dynamics: vehicle accumulation and mobility degradation. Specifically, increases in vehicle activity combined with reductions in average speed result in higher index values, reflecting intensified traffic interactions and congestion conditions. This formulation is therefore particularly useful for line-based sensing systems such as DAS, which capture spatiotemporal variations in aggregated traffic dynamics, particularly under event-driven or disruptive conditions.

To quantify traffic responses to disruptive events, congestion intensity during event periods was compared against matched baseline conditions using a relative change metric. For each analysis window $t$, the percentage change in congestion intensity was defined as

$$\Delta CI\%(t) = \frac{CI_{event}(t) - CI_{baseline}(t)}{CI_{baseline}(t)} \times 100\% \quad (4)$$

where $CI_{event}(t)$ denotes the congestion intensity observed during disruptive events and $CI_{baseline}(t)$ represents the corresponding baseline congestion intensity at the same time. Baseline conditions were defined using non-event days with matched weekday composition to ensure temporal comparability. Specifically, home football gamedays were compared against non-gameday Saturdays without holidays, while the Thanksgiving period was compared against Thursday to Saturday periods without holiday and gameday. All baseline days excluded major events to ensure comparability with the corresponding event windows (Table S1). This event-based comparison is designed to characterize event-associated traffic-state reorganization rather than establish causal effects of individual events, as other external factors may also influence observed traffic dynamics.

## 4 Results

### *4.1 Validation of the proposed DAS traffic-state inference framework*

To evaluate whether the proposed framework can provide robust and computationally efficient traffic-state inference for continuous urban traffic monitoring, we first assessed the performance of three representative deep learning models (YOLOv8, YOLOv11, and RT-DETR) under different training strategies and then compared the resulting traffic volume and speed estimates with those obtained from the conventional slant-stacking approach.

Across all evaluated architectures, model performance improved consistently from synthetic pre-training (Stage 1) to hybrid training (Stage 2), whereas additional fine-tuning using only real data (Stage 3) resulted in only marginal improvements (Fig. 3a). Models trained solely on synthetic data exhibited limited generalization to real DAS observations, with mAP@0.5–0.95 values below 0.35. Incorporating real DAS data during hybrid training substantially improved detection performance across all models, increasing mAP@0.5–0.95 to approximately 0.75–0.77, while maintaining stable convergence. Evaluation on an independent hold-out dataset (Table S2) further confirmed that the hybrid training strategy consistently achieved the highest detection performance. YOLOv8 and YOLOv11 obtained the highest mAP@0.5–0.95 values (0.768), whereas RT-DETR showed slightly lower performance across all training strategies. These results indicate that combining synthetic pre-training with limited real-world annotations provides a robust and scalable strategy for DAS-based traffic inference while substantially reducing the dependence on large manually labeled datasets.

Although quantitative evaluation achieved high detection performance, qualitative analysis identified several challenging scenarios. First, under highly congested conditions, multiple vehicle-induced vibration signatures may overlap and merge into high-energy regions, reducing the separability of individual trajectories and increasing detection uncertainty. Second, simultaneous vehicle movements in opposite directions may generate intersecting DAS trajectories, which can partially obscure individual vehicle signatures and may lead to missed detections or incomplete trajectory extraction. Representative examples of these challenging cases are provided in Fig. S3.

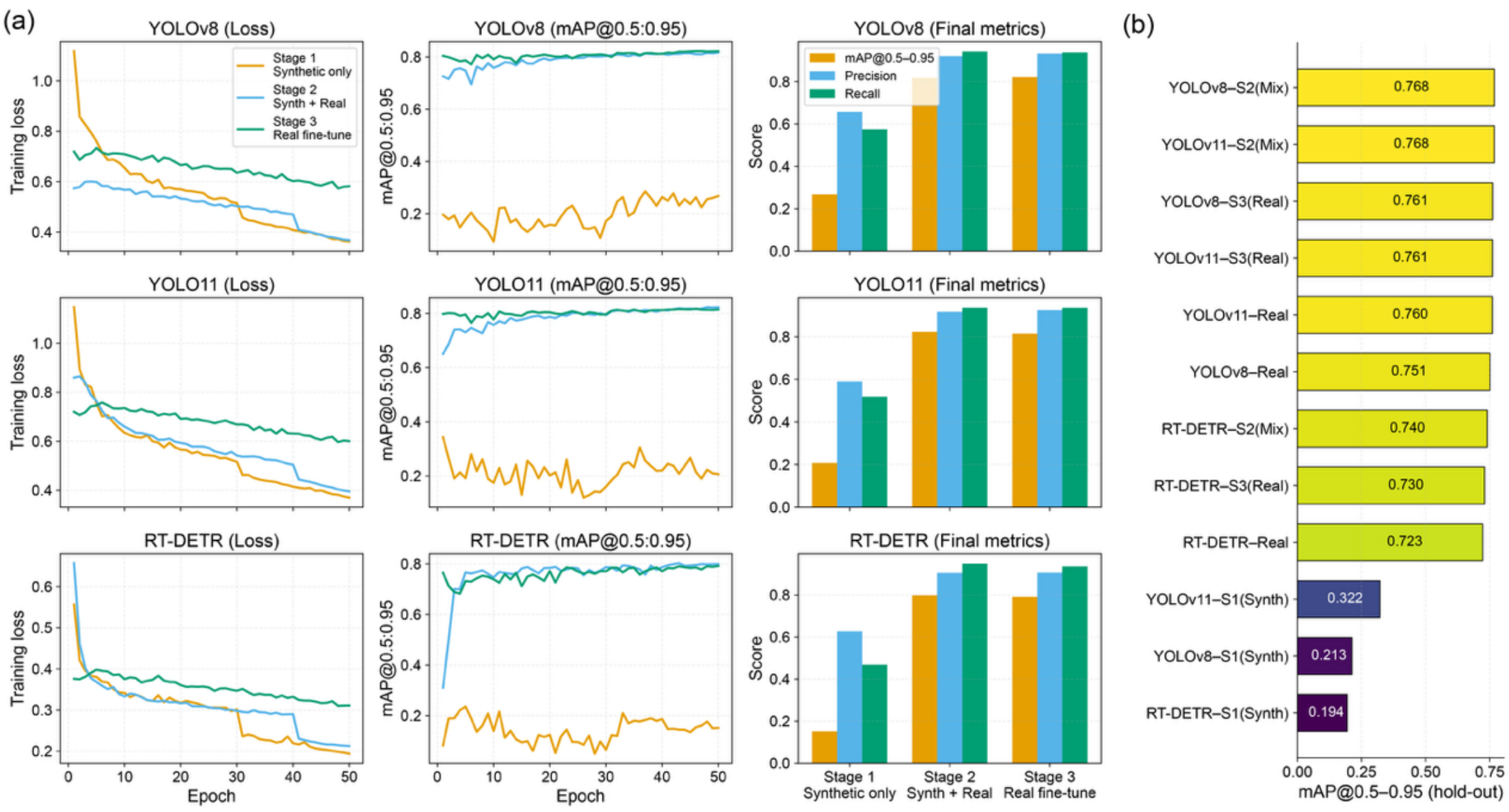

Fig. 3. Training convergence and performance comparison of deep learning-based traffic detection models. (a) Training loss, mAP@0.5–0.95 convergence, and final detection metrics for YOLOv8, YOLOv11, and RT-DETR across the three-stage training scheme. (b) Detection performance on the independent hold-out dataset measured by mAP@0.5–0.95.

Based on the balance between detection accuracy, computational efficiency, and model robustness, the YOLOv8 model trained using the Stage 2 hybrid strategy was selected for subsequent traffic-state inference. Vehicle detections were aggregated to estimate traffic volume and DAS-derived apparent speed and compared with estimates obtained using the conventional 4th-root slant-stacking approach (Fig. 4). Both approaches capture the overall temporal variability in traffic volume, with a clear minimum around 04:00 and two peak periods near 10:00 and 16:00. Quantitatively, the deep learning-based estimates show stronger agreement with manual annotations, as reflected by higher correlation and lower absolute error metrics ($r = 0.931$, $R^2 = 0.866$, MAE = 0.608), compared with slant stacking ($r = 0.882$, $R^2 = 0.779$, MAE = 0.798). This difference is likely related to the sensitivity of slant stacking to signal energy and peak stability, as illustrated by the representative examples in Fig. 4c and 4d, where low-energy or unstable vehicle signals are more prone to missed detections. Notably, analysis of month-long data from November 2023 further corroborates this pattern, indicating that under high-volume daytime conditions, slant-stacking exhibits a systematic downward bias relative to YOLO-based estimates (Fig. S4). Fig. 4b compares average apparent traffic speeds estimated using the deep learning-based framework and the slant-stacking approach. Although the two estimates show a statistically significant Spearman correlation ($\rho = 0.278$, $p = 7.46 \times 10^{-9}$), this comparison indicates that the two DAS-based methods capture related but not identical apparent speed patterns. The discrepancy becomes more pronounced under congested conditions, where slant stacking tends to produce higher apparent speed estimates (Fig S5). This difference may result from variations in signal representation, trajectory extraction mechanisms, and sensitivity to signal overlap under complex traffic conditions.

In addition to improved inference performance, the proposed framework substantially increased computational efficiency. Under the same computational environment, the inference time was reduced from 1201 s using the conventional slant-stacking workflow to approximately 23 s using the deep learning-based framework, representing an approximately 50-fold improvement (Fig. 4e). This comparison considers only the inference time of the trained model and excludes the one-time costs of manual annotation and model training. This improvement enables efficient processing of continuously acquired DAS observations and provides a practical foundation for near-real-time traffic-state monitoring. Collectively, these results demonstrate that the proposed framework provides sufficiently robust, accurate, and computationally efficient traffic-state inference to support the subsequent analyses of event-driven urban traffic dynamics.

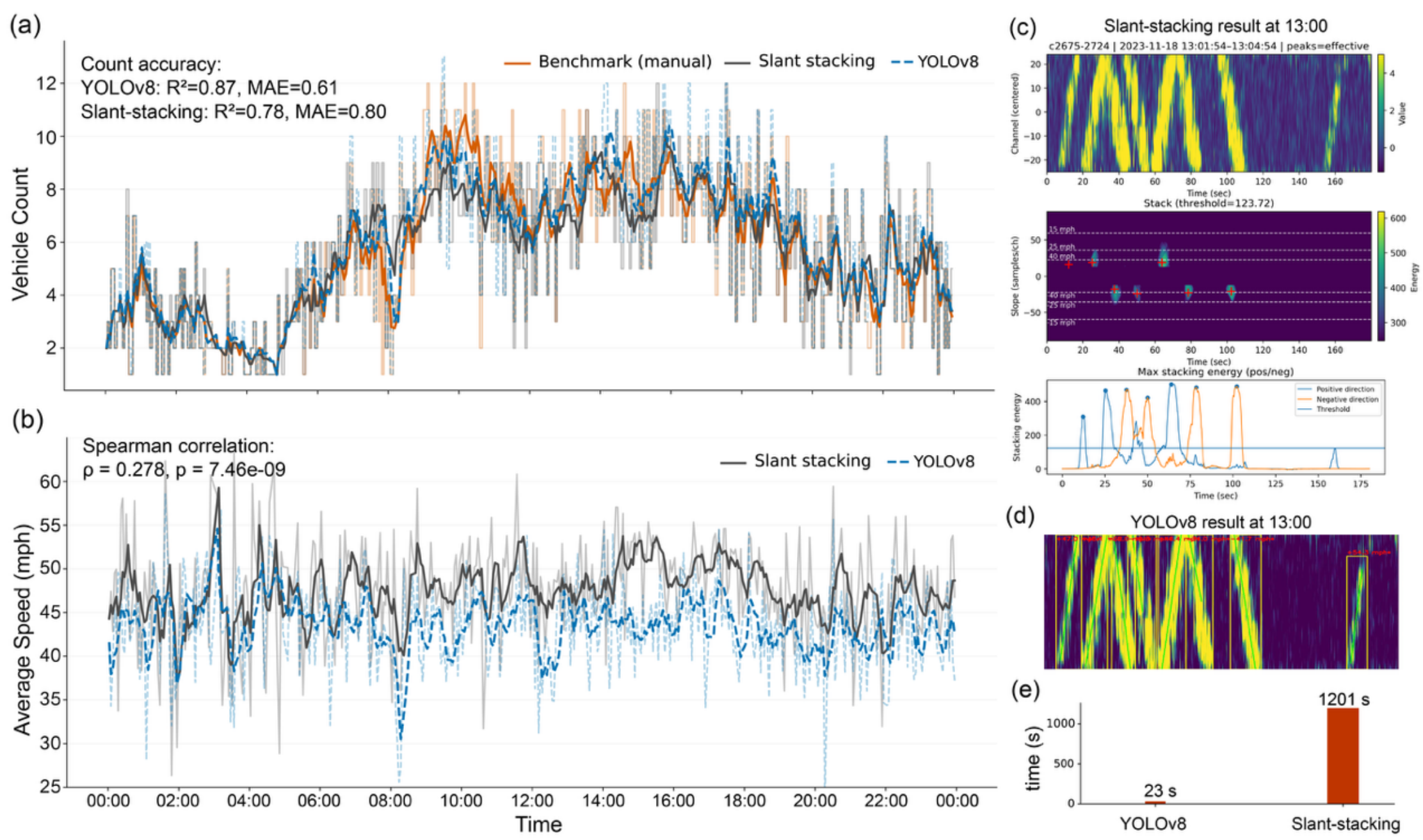

Fig. 4. Comparison of vehicle volume and apparent speed estimation using deep learning and slant-stacking methods. (a) Time series of detected vehicle volume, showing raw 3-min records (bars) and 15-min aggregated trends (lines), benchmarked against manual annotations. (b) Time series of estimated average apparent speed, showing raw 3-min records (bars) and 15-min aggregated trends (lines). (c) Example of slant-stacking output and peak detection used for vehicle signature extraction. (d) Example of deep learning-based detection of vehicle-induced trajectories in a DAS spatiotemporal image. (e) Comparison of computational efficiency between deep learning-based inference and slant stacking.

### *4.2 Event-scale traffic state dynamics and urban activity inference*

Using the validated DAS traffic-state inference framework, we next investigate how different disruptive events reshape urban traffic dynamics over time. Fig. 5 summarizes the temporal evolution of traffic volume, average speed, and congestion intensity from October to November 2023, revealing distinct patterns of mobility reorganization associated with home football gamedays and the Thanksgiving holiday.

Under non-event conditions, traffic volume exhibits highly stable and repeatable diurnal rhythms of the area's routine commute-driven mobility, characterized by low nighttime activity, a gradual morning buildup, and elevated daytime levels. In contrast, event periods are associated with distinct departures from this baseline, including temporally concentrated surges during home football gamedays and broader suppressions during the Thanksgiving holiday. In contrast to traffic volume, average speed remains comparatively stable, typically ranging from approximately 45 to 50 mph, with only modest reductions during home football gamedays and slight increases during the Thanksgiving holiday. By integrating traffic volume and speed, the congestion intensity indicator captures congestion-induced friction in traffic flow, producing clearer and more interpretable signatures of event-driven traffic states than either variable alone.

To further quantify event-driven traffic state changes, we compare diurnal variations in congestion intensity during event periods against matched non-event baselines. For a representative football gameday (Texas A&M Aggies vs. South Carolina Gamecocks) on Oct 28, 2023, congestion intensity consistently exceeds the baseline mean across most daytime hours and frequently falls outside the baseline variability envelope (±1 SD). When averaged over the full diurnal cycle, congestion intensity on the representative gameday increases by 34.2% relative to baseline Saturdays, indicating a substantial system-level increase in traffic pressure. The temporal evolution of congestion intensity reveals a distinct intra-day structure in the gameday response. Under both baseline and event conditions, congestion intensity reaches its minimum during the early morning hours around 05:00. However, congestion levels on the gameday are elevated by approximately 30–40% relative to baseline during the pre-dawn period (before 05:00), likely reflecting increased nighttime mobility associated with event-related activities preceding the game day, such as the campus tradition known as the Midnight Yell Practice, typically held around midnight. Following this early-morning minimum, congestion intensity on the gameday increases more rapidly than under baseline conditions and exhibits multiple localized peaks prior to the scheduled kickoff time (11:00), with distinct peaks observed during the early morning (approximately 06:00–07:00) and again near 10:00. These pre-event peaks are consistent with heterogeneous arrival patterns, whereby non-local attendees tend to arrive earlier in the day, while local residents typically access the event area closer to kickoff. The largest congestion anomaly occurs around 20:00, when congestion intensity nearly doubles relative to baseline, likely reflecting elevated post-event mobility associated with celebrations and subsequent social activities following a home victory. Together, these patterns illustrate how social events can reorganize urban traffic rhythms across the full diurnal cycle, extending well beyond the event window itself.

In contrast, congestion intensity during Thanksgiving exhibits a significant reduction relative to matched non-event baselines. As shown in Fig. 5, diurnal congestion intensity values consistently remain below the baseline mean and largely below the lower bound of baseline variability. When averaged over the full diurnal cycle, congestion intensity during the Thanksgiving period decreases by 49.9% relative to baseline conditions, reflecting a substantial reduction in overall traffic activity. Unlike football gamedays, which concentrate travel into distinct temporal peaks, Thanksgiving produced a sustained flattening of the diurnal traffic profile. This pattern reflects the temporary disruption of work-, school-, and other institution-based activities, which weakens the temporal synchronization of everyday travel and reduces mandatory peak-period demand. Consequently, the transportation impacts of major public holidays are manifested not as localized congestion amplification but as a broad reorganization of routine mobility across the urban transportation system.

Taken together, these contrasting traffic-state responses demonstrate that disruptive events influence urban transportation systems through fundamentally different mechanisms. Rather than simply increasing or decreasing traffic demand, different event types reorganize the timing, intensity, and synchronization of urban mobility, producing distinct traffic-state evolution that can only be fully characterized through continuous spatiotemporal observations.

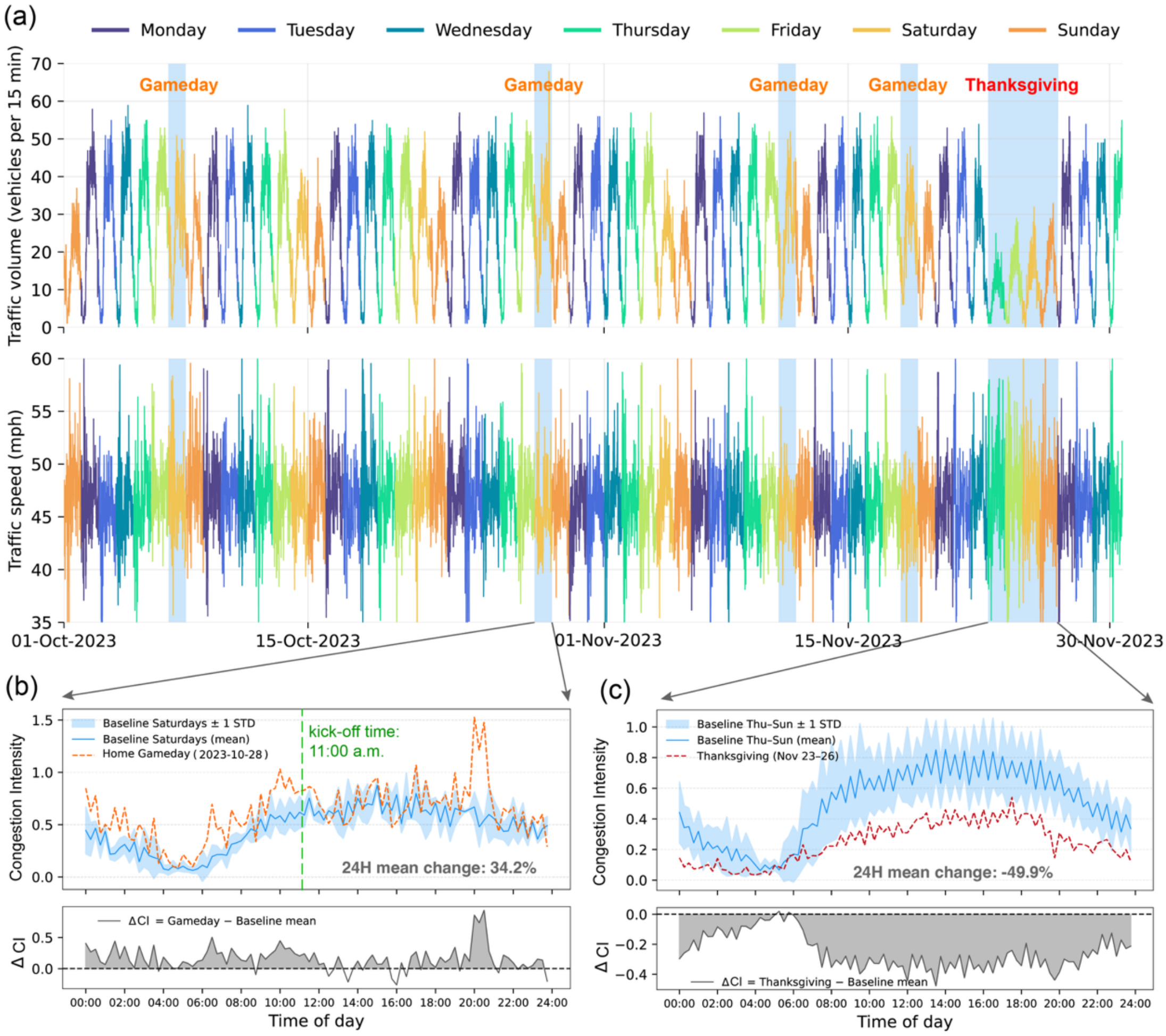

Fig. 5. Long-term variations in traffic patterns during home football gamedays and the Thanksgiving holiday. (a) Traffic volume and average speed time series show regular weekday variability and longer-term fluctuations, with shaded intervals indicating home football gamedays and the Thanksgiving holiday. Bottom panels compare diurnal congestion intensity for (b) a representative gameday and (c) the Thanksgiving holiday with matched baseline conditions (mean ± 1 SD). Traffic statuses are summarized using congestion intensity index, defined as traffic volume divided by average speed, to characterize temporal variations in traffic states across different event types.

### *4.3 Spatial heterogeneity of event-driven traffic responses*

Traffic responses to disruptive events are rarely spatially uniform. Instead, the magnitude and timing of congestion intensity are conditioned by roadway function, network position, and the spatial organization of travel demand. Leveraging the continuous corridor-scale observations provided by DAS, we examine how different roadway segments respond to football gamedays and the Thanksgiving holiday, thereby revealing the spatial heterogeneity of event-driven traffic dynamics across the urban transportation network (home football gamedays in Fig. 6 and the Thanksgiving holiday in Fig. 7).

Fig. 6a shows the 24-hour average percentage change in congestion intensity relative to matched baseline conditions during a representative football gameday (Texas A&M Aggies vs. South Carolina Gamecocks). Congestion intensity increases are observed across all roadway segments during the gameday, with considerable inter-segment variability ranging from roughly 10% to 65% relative to baseline. Among all segments, the primary arterial along F&B Road exhibits the largest increase in congestion intensity, with average changes consistently exceeding 50%. This segment functions as a main access corridor for spectators traveling toward Kyle Field, thereby concentrating on a substantial portion of event-related traffic demand. In contrast, secondary roadway segments leading toward internal campus facilities and recreational areas show more modest congestion increases, indicating weaker sensitivity to gameday-related travel. The segment-specific diurnal congestion profiles shown in Fig. 6b–d further demonstrate that this spatial heterogeneity persists throughout the day. For high-impact segments (Fig. 6b), the sharp increases in congestion intensity appear closely associated with event-related human activities. Multiple nearly doubled congestion surges are observed in the early morning hours (around 00:00), as well as around 07:00 and 10:00, consistent with pre-event social gatherings on the previous evening and successive waves of inbound traffic as spectators begin traveling toward the stadium. By comparison, low-impact segments (Fig. 6c–d) do not exhibit the pre-event congestion buildup observed along the primary stadium access corridor and instead show reduced congestion intensity relative to baseline conditions. This contrasting pattern may reflect the redistribution of traffic induced by gameday traffic control. Access restrictions and traffic management around the stadium may limit vehicle movements along particular road segments and redirect event-related traffic toward designated access routes, thereby concentrating traffic on some segments while reducing it on others.

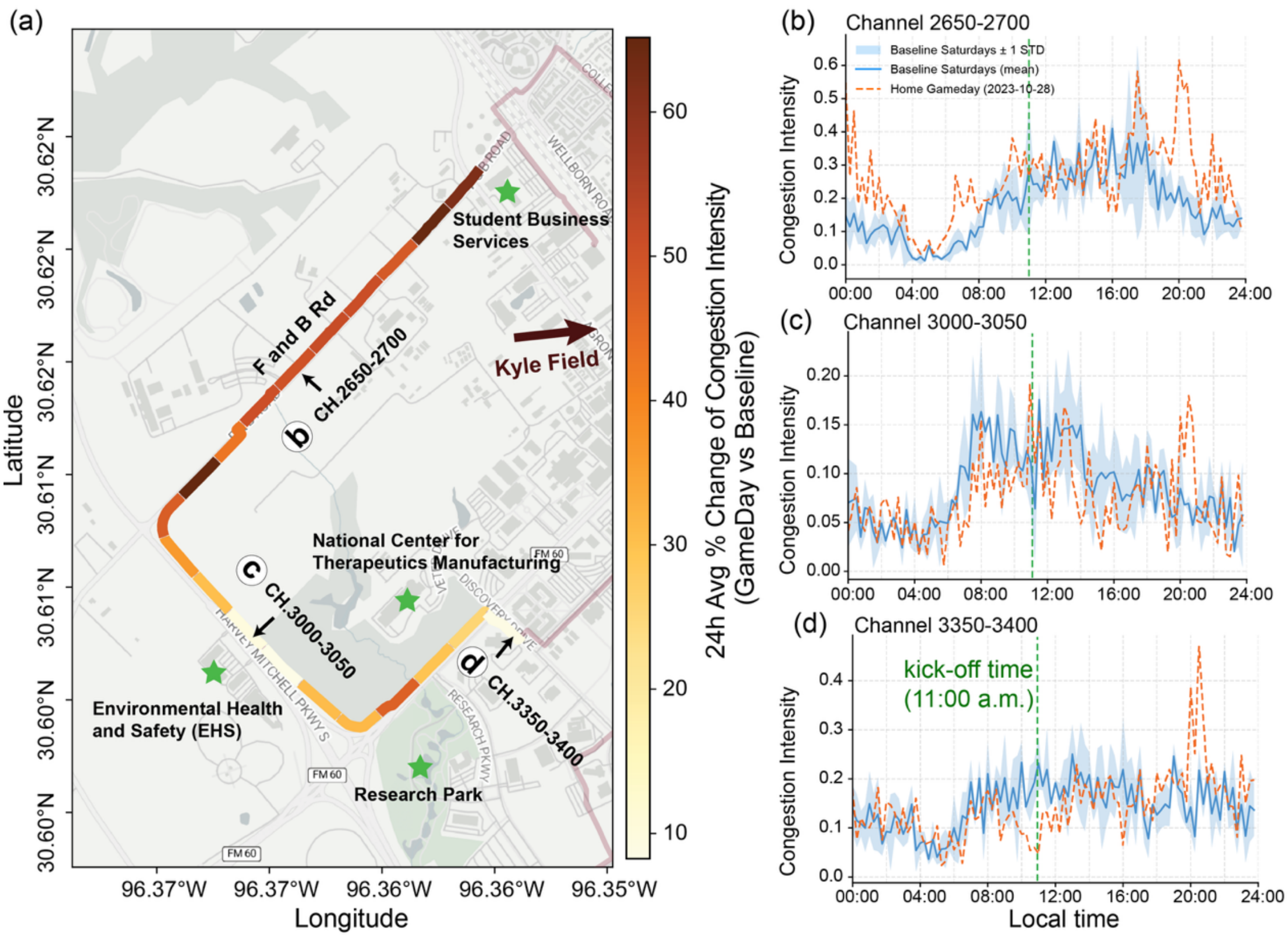

Fig. 6. Spatial variation in congestion intensity responses to a football gameday along different DAS-monitored roadway segments, shown by (a) 24-hour average congestion intensity change relative to baseline and (b–d) segment-specific diurnal congestion profiles.

While congestion intensity is broadly reduced during the Thanksgiving holiday, spatial heterogeneity persists across roadway segments, indicating unevenly distributed traffic suppression across the urban road network (Fig. 7). Figure 7a presents the 24-hour average percentage change in congestion intensity relative to baseline conditions, showing reductions of approximately 20% to over 50% across all monitored segments, with the strongest suppression concentrated on high-volume arterial segments along F&B Road. In contrast, lower-volume secondary segments exhibit more moderate reductions. Temporally, Thanksgiving substantially alters the diurnal patterns of traffic, particularly on high-volume arterials. Under baseline conditions, congestion intensity reaches a minimum around 05:00, followed by a sharp increase between 07:00 and 08:00, corresponding to the morning commuting peak. During Thanksgiving, this early-morning surge is strongly attenuated, with peak congestion intensity reduced by approximately 30–40%, resulting in a flattened diurnal profile in which the morning peak is largely absent, and congestion intensity remains persistently low throughout daytime hours (Fig. 7b). In contrast, secondary and lower-volume segments display a distinct temporal response, with traffic reductions primarily confined to daytime periods (roughly 10–20%), while early-morning (before 06:00) and evening (after 18:00) traffic levels remain broadly comparable to baseline conditions (Fig. 7c–d). Together, these results indicate that event-driven traffic responses are not spatially uniform but are strongly conditioned by roadway function and baseline traffic demand.

Taken together, these findings demonstrate that event-driven traffic responses are structured by roadway hierarchy rather than being uniformly distributed across the urban transportation network. High-capacity arterial corridors primarily accommodate regionally coordinated travel associated with major events and routine commuting, whereas secondary roadways maintain a greater proportion of localized mobility. Consequently, disruptive events reorganize traffic systems unevenly across urban space, highlighting the importance of considering roadway function and location when evaluating transportation resilience and planning event-specific traffic management strategies.

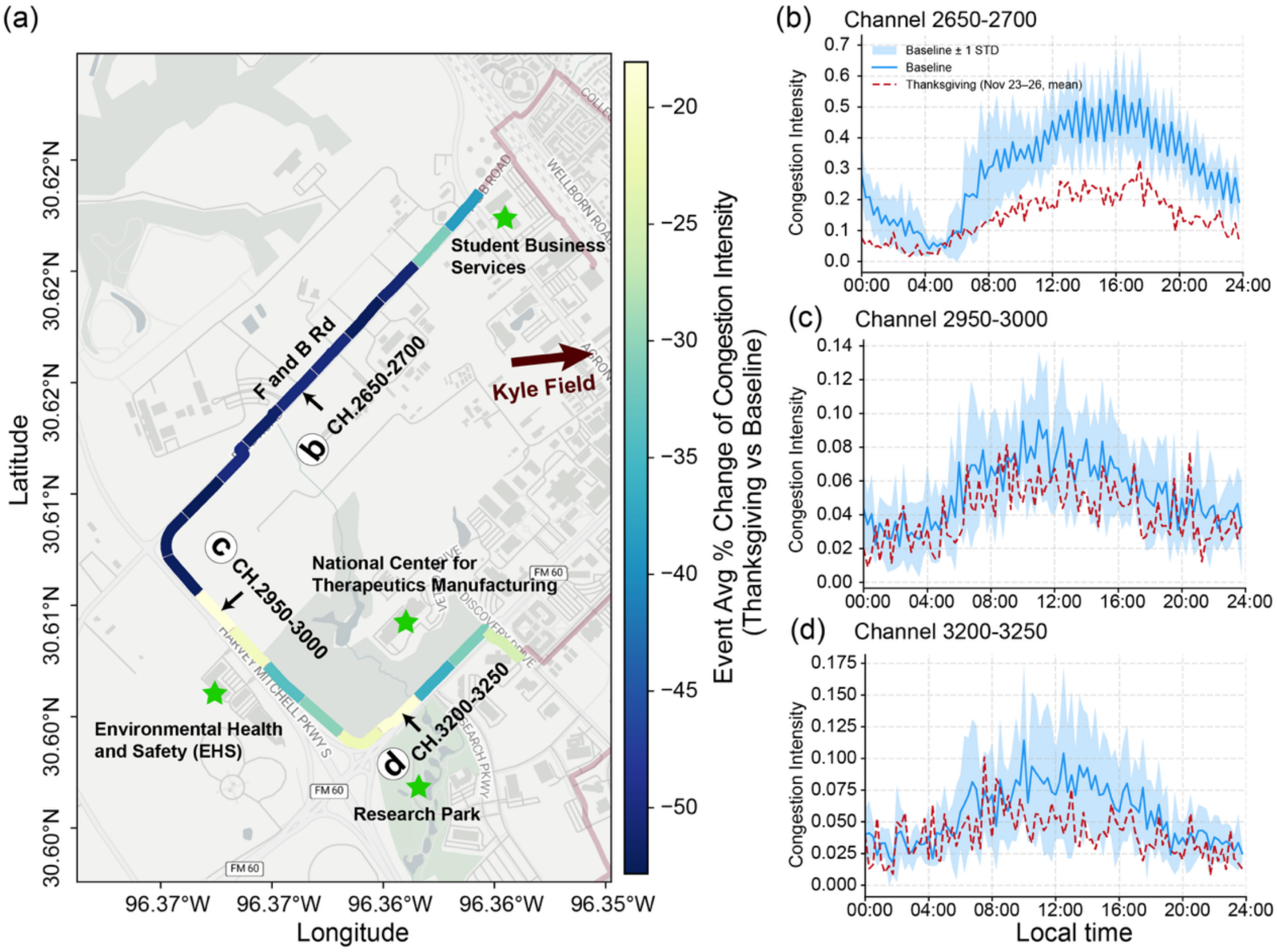

Fig. 7. Spatially varying congestion intensity reductions during the Thanksgiving holiday across different DAS-monitored roadway segments, illustrated by (a) event-averaged congestion change relative to baseline and (b–d) segment-specific diurnal congestion patterns.

## 5. Discussion

### *5.1 Deep learning-enabled DAS for urban traffic monitoring*

The integration of deep learning and distributed acoustic sensing (DAS) technology offers a promising approach for scalable and privacy-preserving traffic monitoring. In this study, we develop a deep learning-enabled DAS analytical framework for vehicle trajectory detection and traffic state inference from continuous ground vibration signals. Compared with conventional traffic sensing approaches, DAS provides a fundamentally different sensing paradigm. Unlike camera-based systems, it does not rely on visual data and therefore avoids privacy concerns associated with image acquisition (Gao et al., 2025). In contrast to GPS or mobile phone data, which are often subject to sampling bias and participation constraints (Herrera et al., 2010; Y. Wang et al., 2025), DAS enables continuous, infrastructure-based sensing with high temporal resolution and minimal dependence on user behavior. These characteristics make DAS particularly well-suited for monitoring traffic dynamics in complex or disruptive conditions, where traditional sensing systems may be sparse, incomplete, or unreliable (Deng et al., 2025). Earlier DAS-based traffic studies have largely depended on classical signal-processing workflows, such as slant stacking (Wang et al., 2021), matched template algorithm (Lindsey et al., 2020), and related trajectory-enhancement procedures (Yuan et al., 2024), to identify vehicle-induced wavefields and estimate speed or flow. In practice, roadway DAS data are often affected by weak amplitudes and strong environmental noise, all of which make purely signal-based extraction less robust under realistic urban traffic conditions (Deng et al., 2025).

Against this background, the main contribution of our framework is not simply to replace one detector with another, but to shift the inference paradigm from explicit signal processing to deep learning-based pattern recognition. Our results reveal that slant stacking is effective when vehicle wavefields are relatively coherent and separable, but its performance can decrease when low-energy trajectories are partially obscured. Our deep learning-based detector can exploit shape, continuity, local contrast, and contextual cues simultaneously, allowing it to recognize trajectories with weak energy signatures. This shift in paradigm enables the deep learning-based approach to achieve substantial improvements in vehicle detection accuracy, thereby providing a robust foundation for continuous traffic-state inference rather than vehicle detection alone. Despite these improvements, the qualitative failure analysis indicates that the proposed detector remains challenged under highly congested conditions and simultaneous bidirectional vehicle movements. When multiple vehicle-induced signatures overlap or intersect, individual trajectories may become difficult to separate, resulting in increased detection uncertainty or missed detections. Future work could address these challenges through direction-aware trajectory modeling and temporal tracking, which may better exploit directional and temporal continuity to distinguish overlapping or intersecting vehicle signatures.

However, beyond traffic volume detection, we observe notable differences in speed estimation between the deep learning-based approach and the traditional slant-stacking method, particularly under high-traffic conditions. Specifically, speeds derived from slopes fitted to deep learning-detected vehicle trajectories tend to be lower than those estimated using the traditional slant-stacking approach. This deviation is likely attributable to fundamental differences in the algorithmic principles underlying the two methods. The slant-stacking approach estimates speed by identifying energy maxima in the spatiotemporal wavefield, which are often dominated by faster or more energetic vehicles when multiple vehicles overlap (Yilmaz, 2001). As a result, under congested conditions, this energy aggregation effect can bias the estimated speed upward. In contrast, the deep learning-based approach identifies individual vehicle trajectories and estimates speed from trajectory-level slope fitting (e.g., PCA-based methods), effectively capturing a broader distribution of vehicle behaviors. This leads to more conservative and potentially more representative average speed estimates under complex traffic conditions.

Beyond accuracy, computational efficiency is equally important for the future of DAS-enabled real-time traffic monitoring. DAS systems inherently provide continuous, high-frequency measurements of ground vibrations, typically at sampling rates of tens to hundreds of Hz, enabling sub-second and near-real-time data acquisition along fiber-optic infrastructure (A. Hartog, 2017). This sensing advantage is only useful for real-time traffic monitoring if downstream processing is equally efficient. Traditional slant-stacking workflows typically require repeated scanning over candidate slopes, which can become computationally expensive for long fiber segments or continuous monitoring. In contrast, the approximately 50-fold speedup observed in this study is not only substantial in magnitude but also critical from a system-level perspective. It indicates that deep learning-based approaches can effectively align high-frequency data acquisition with efficient inference, thereby enabling near-real-time interpretation of continuously acquired DAS data. This alignment is essential for translating DAS from a passive sensing modality into a real-time traffic monitoring system.

Nevertheless, our findings do not support the conclusion that deep learning algorithms can completely replace traditional signal processing methods. Conventional methods still offer several key advantages, including a transparent physical foundation, interpretable outputs, and the ability to operate with little or no reliance on labeled data during implementation (Wang et al., 2022). Deep learning-based methods, by contrast, offer superior robustness and scalability once an effective training pipeline is available, but it introduces dependencies on labeled data, model transferability, and deployment-specific calibration. For this reason, a more productive perspective is to view deep learning-based methods and conventional signal-processing approaches as complementary rather than strictly competing paradigms. Signal-processing methods remain valuable for physical interpretability, weakly supervised inference, and benchmarking, whereas deep learning is particularly advantageous for high-congestion, noisy, and large-scale monitoring scenarios where handcrafted procedures become difficult to maintain or scale. Together, these findings suggest that deep learning and conventional signal-processing approaches should be viewed as complementary components of future DAS-enabled traffic monitoring systems rather than competing alternatives.

### *5.2 Mobility reorganization under disruptive events*

Urban traffic systems rarely respond uniformly to disruptive events. Instead, the impacts of different disruptions depend on how human activities are reorganized across space and time, producing distinct patterns of mobility reorganization rather than simple changes in traffic volume (Donovan & Work, 2017; Feng et al., 2025). This perspective is consistent with activity-based travel theory, which emphasizes that changes in daily activity schedules reshape both the temporal organization and spatial distribution of travel demand (Hägerstrand, 1970). Home football gamedays and the Thanksgiving holiday provide two representative examples of these contrasting forms of mobility reorganization: the former corresponds to planned, destination-oriented events that concentrate travel demand, whereas the latter reflects temporary interruptions of routine activity systems.

Planned special events, represented here by home football gamedays, generate a characteristic mobility amplification pattern. This pattern reflects the temporal redistribution of travel demand commonly associated with planned special events, where mobility is structured by coordinated arrival times, staged participation, and synchronized departure behavior. Similar multi-peak dynamics have been widely documented in stadium events and large-scale gatherings, where congestion patterns emerge not only from increased demand but also from temporal clustering of activity schedules (Bassolas et al., 2020; Polson & Sokolov, 2017). From an activity-based perspective, these patterns reflect the decomposition of typical daily activity chains into event-centered travel episodes, leading to temporally concentrated but highly structured traffic fluctuations.

Major public holidays, represented here by Thanksgiving, exhibit a contrasting mobility suppression pattern characterized by sustained reductions in congestion intensity and a flattened diurnal traffic profile. Rather than uniformly intensifying localized congestion, holiday periods temporarily disrupt routine work-, school-, and other institution-based activities, reducing mandatory peak-period travel and increasing the relative prominence of discretionary and locally oriented trips; congestion may nevertheless shift toward leisure destinations, commercial centers, and intercity corridors (Yu et al., 2024; Zhou et al., 2025). From an activity-based travel perspective, this response reflects a temporary reorganization of daily activity schedules, leading to weaker temporal synchronization of urban mobility and consequently less pronounced commuting peaks.

Taken together, these findings suggest that disruptive events should not be interpreted simply as increases or decreases in traffic demand. Instead, different event types reorganize urban mobility by altering the timing, destinations, and synchronization of travel activities. Capturing these evolving traffic states therefore requires observation frameworks capable of continuously monitoring traffic dynamics across both space and time, rather than relying solely on aggregate traffic counts or isolated point measurements.

### *5.3 DAS as a continuous urban traffic observatory*

The findings of this study suggest that the significance of DAS extends beyond its sensing capability. More fundamentally, DAS enables a shift in how urban traffic systems are observed, from fragmented measurements collected at isolated locations to continuous observations of traffic-state evolution along transportation corridors. Conventional traffic monitoring infrastructures, including loop detectors, traffic cameras, and probe vehicle data, have substantially advanced transportation research and traffic operations, but they remain constrained by either discrete spatial sampling or incomplete population coverage (Xing et al., 2022). As a result, they often provide spatially discontinuous observations of traffic conditions rather than continuous observations of how traffic states evolve along transportation corridors. By contrast, DAS transforms existing fiber-optic infrastructure into a dense linear sensor array that senses ground vibrations at meter-scale spatial and second-level temporal resolution, enabling traffic-state evolution to be tracked continuously along entire corridors rather than at isolated points (Lindsey & Martin, 2021; Zhan, 2019). This capability enables traffic systems to be interpreted as continuously evolving spatiotemporal processes rather than collections of independent traffic measurements, providing new opportunities to investigate mobility dynamics, congestion propagation, and network responses to disruptive events. This continuous coverage, however, comes at the cost of spatial granularity: the reported traffic-state indicators are aggregated over corridor windows rather than resolved at the meter scale, so variations within a window are not individually captured, though finer-grained estimation becomes feasible under higher traffic volumes or longer observation windows.

From a transportation systems perspective, the primary value of DAS does not necessarily lie in achieving higher point-level detection accuracy than mature sensing technologies such as cameras or loop detectors. Indeed, under well-controlled environments, camera-based systems remain highly accurate for vehicle detection and classification. Instead, DAS offers a complementary sensing paradigm with distinct advantages in scenarios where continuous spatial coverage, privacy preservation, and infrastructure resilience are of primary importance. Because DAS passively measures ground vibrations rather than collecting visual or personal location information, it substantially reduces the privacy concerns associated with image-based monitoring, such as the capture of faces or license plates, while remaining independent of the user participation required by probe-based data sources (Lindsey et al., 2020). Moreover, unlike vision-based systems whose performance may deteriorate under poor lighting, adverse weather, or power disruptions, DAS continuously records vibration signals along the fiber network, making it particularly suitable for long-term monitoring and disruptive-event observations (Deng et al., 2025). These characteristics make DAS particularly valuable for investigating traffic dynamics under planned events, emergency situations, and infrastructure disruptions where maintaining continuous situational awareness is critical for transportation system management and resilience.

Taken together, DAS should not be regarded as a universal replacement for existing traffic sensing technologies. Rather, its greatest practical value lies in strategic deployment along critical transportation corridors, highways, tunnels, and urban areas vulnerable to traffic disruptions, where continuous observations provide information that conventional point-based sensors cannot readily obtain. Nevertheless, the practical implementation of DAS depends on the availability of suitable fiber-optic infrastructure and the costs associated with DAS interrogator deployment and operation. Therefore, at the current stage of technological development, DAS is more appropriately viewed as a complementary sensing technology for targeted deployment rather than a universal monitoring solution. Importantly, extensive fiber-optic infrastructure has already been deployed in many urban areas but remains underutilized for transportation monitoring. Repurposing these existing assets provides a promising pathway for future DAS applications. As DAS hardware continues to mature, data processing becomes more efficient, and economies of scale are achieved, these implementation barriers may gradually be reduced, facilitating broader adoption of DAS-enabled traffic monitoring.

### *5.4 Limitations and future work*

Several limitations should nevertheless be acknowledged. First, this study is based on a single DAS deployment near Texas A&M University. Although the proposed framework demonstrates robust performance under the investigated conditions, its generalizability may vary across different traffic environments, such as dense downtown areas or highways, due to differences in roadway characteristics, fiber-ground coupling conditions, background vibration patterns, and traffic compositions. In addition, the estimated apparent speed has not been validated against independent reference measurements, such as radar, GPS trajectories, or camera-based observations. Future validation across diverse DAS deployments and with independent traffic measurements will be necessary to further evaluate the transferability of the framework and the reliability of DAS-derived apparent speed estimation. Second, the proposed congestion intensity index should be interpreted as a relative congestion indicator rather than a direct measurement of physical traffic density. Since DAS captures aggregated vehicle signatures along a sensing line, it does not directly provide lane-level occupancy or roadway geometry information required for estimating physical density. Third, the event-based analysis in this study is designed to characterize event-associated traffic-state dynamics rather than establish causal relationships. Although matched baseline periods were used to reduce variability associated with regular weekly mobility patterns, other contextual factors, including weather conditions, academic schedules, and concurrent campus activities, may also contribute to observed traffic variations. Future research should integrate additional contextual variables within a statistical modeling framework to better quantify the independent contributions of different disruptive events.

Future research may extend this framework in several directions. One promising direction is to integrate DAS with complementary data sources, such as loop detectors or GPS-based mobility indicators, to develop multimodal traffic monitoring systems. Another direction is to reduce reliance on manual labeling through semi-supervised, self-supervised, or adaptive learning strategies, which may further improve the transferability of the proposed framework across different DAS deployments while lowering the barrier to large-scale deployment.

## 6. Conclusion

This study demonstrates that integrating distributed acoustic sensing (DAS) with deep learning provides an efficient framework for continuous traffic-state inference from ground-vibration observations. The proposed framework improves the robustness of vehicle detection under noisy and congested conditions while substantially reducing computational cost compared with conventional signal-processing approaches. Beyond vehicle-level detection, the framework enables continuous monitoring of urban traffic states from aggregated DAS observations.

Using home football gamedays and the Thanksgiving holiday as representative disruptive events, we show that different event types reorganize urban mobility in distinct ways, producing characteristic temporal patterns and spatially heterogeneous roadway responses. These findings highlight the potential of DAS to capture event-driven traffic dynamics that are difficult to observe using conventional point-based sensing systems. More broadly, this study suggests that DAS should be viewed not simply as a new sensing technology, but as a complementary infrastructure for continuous observation of urban traffic systems. By supporting continuous and privacy-preserving monitoring of traffic-state evolution, DAS has the potential to complement existing traffic sensing technologies and contribute to future research on urban mobility, transportation resilience, and spatially heterogeneous traffic dynamics. Future work should improve model transferability across sensing environments and further evaluate the applicability of DAS under a broader range of roadway and disruptive-event scenarios.

**Declaration of competing interest**

No conflict of interest exists in the publication or creation of this manuscript. The manuscript has been reviewed and approved by all co-authors for publication. We certify that the manuscript is not under review for publication by any other journals.